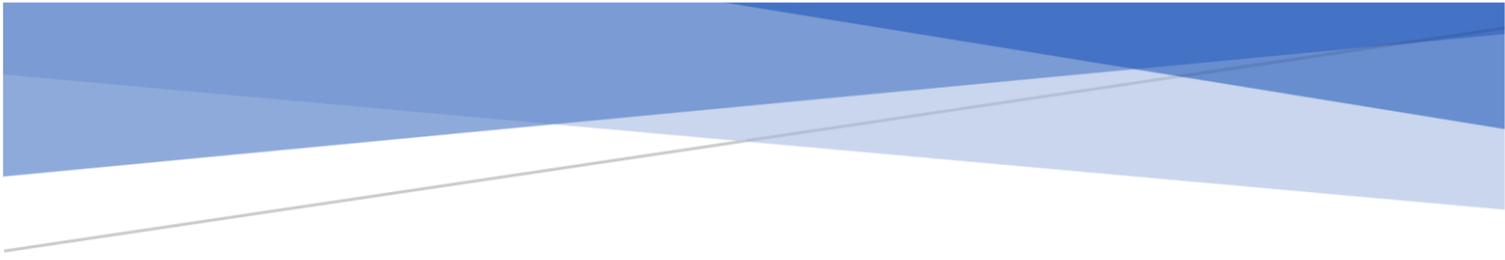

# MODEL SELECTION WITH GRAPHICAL NEIGHBOUR INFORMATION

Robert O'Shea
robert.o'shea.17@ucl.ac.uk

# Contents

# Introduction

Accurate model selection is a fundamental requirement for statistical analysis (1–5). In many real-world applications of graphical modelling, correct model structure identification is the ultimate objective. Standard model validation procedures such as information theoretic scores and cross validation have demonstrated poor performance when $n \ll p$. Specialised methods such as EBIC, StARS and RIC have been developed for the explicit purpose of high-dimensional Gaussian graphical model selection. We present a novel model score criterion, Graphical Neighbour Information. This method demonstrates oracle performance in high-dimensional model selection, outperforming the current state-of-the-art in our simulations. The Graphical Neighbour Information criterion has the additional advantage of efficient, closed-form computability, sparing the costly inference of multiple models on data subsamples. We provide a theoretic analysis of the method and benchmark simulations versus the current state of the art.

## Description of the problem

Graphical models have demonstrated utility in various fields of high dimensional statistical analysis. This modality of analysis allows the inference of interaction networks from multidimensional datasets. In such networks, variables are represented graphically as vertices and vertices correspond to dependencies between their incident vertices. Graphical models provide a concise representation of probabilistic dependencies between variables. Complex systems of interactions between pairs of variables may be inferred from observational datasets, under certain conditions. This paradigm has demonstrated broad applicability across various fields, including genomics, imaging and structural engineering. A major problem in the field of high dimensional graphical inference is structure selection and validation. Currently available methods suffer from poor accuracy or high computational complexity. We present an algorithm which quantifies the information of a graph structure on a joint probability distribution. We demonstrate that this measure may be used to select the optimally fitting graph from a set of candidate graphs. We perform a benchmark simulation of model selection in the high dimensional setting and demonstrate that of model selection criterion outperforms the current state of the art in the high dimensional setting, matching it in the low dimensional setting. We provide a computational complexity analysis of our criterion, demonstrating that it is of the order $O(p)$ in the dimension.

Under the gaussian assumption, dependencies between variables may be inferred from the inverse covariance matrix of the joint probability distribution of the data. The sparsity pattern of the inverse gaussian covariance matrix encodes the structure of the graph, such that non-zero entries between variables correspond to incident edges. Therefore, the task of graph inference is represented as a task of inverse covariance matrix estimation. In the high dimensional setting, the maximum likelihood estimate of the inverse covariance matrix is known to be unstable.

L1-regularised approaches to gaussian graphical inference such as the graphical lasso (4) have facilitated graphical modelling by assuming *sparsity* of the true graph. Under this assumption, highly complex models are penalised according to a manually selected lambda parameter. Due in part to the difficulty of regularisation optimisation, several tuning free graph inference approaches have been developed, such as TIGER (6), the de-sparsified graphical lasso (7,8) and the false positive rate controlled lasso(9).

Graphical Model Selection in the high dimensional setting has proven to be a challenging task. The shortcomings of classical approaches such as cross validation and standard information theoretic model scores have been addressed with several specialised graphical model selection techniques. We discuss the deficiencies of the current state of the art in the following paper. We present a novel model selection criterion, Graphical Neighbour Information. This method provides a concise solution to the problem of graphical model selection, matching the performance of the oracle in our high-dimensional simulations. As this method is a model score, it spares the computationally demanding process of multiple model inferences required by methods such as StARS and cross-validation. The method has the additional advantages of full dataset utilisation and tuning free deployment.

## Current state of the art in graphical model selection

Several approaches have been developed to empirically estimate the optimal lambda parameter for a given dataset. Under cross validation, the precision matrix is estimated from a training sample of the dataset. A log-likelihood loss function may subsequently be calculated between the estimated precision matrix and the maximum likelihood estimate of the test sample covariance matrix. Although this approach enjoys sound statistical properties in the low dimensional setting, it has been shown to select complex models in the high dimensional setting. Such models tend to overfit the data (10). Furthermore, overly complex models provide little insight into the underlying generative process. Information theoretic methods such as the Akaike Information Criterion and the Bayesian information criterion suffer from similar overfitting in the case that the number of variables exceeds the number of samples (11).

Consequently, several model validation approaches have been developed specifically for use in graphical modelling. The Extended Bayesian Information Criterion (12) addresses the overfitting problem encountered by information theoretic approaches by applying an additional penalty to the number of parameters which is proportional to $p$. This method shows improved performance over the typical Bayesian Information criterion (12). The validity of the score depends on the appropriate selection of a gamma tuning parameter. This method has the advantage that it is requires a single estimate of the model. We demonstrate in simulations that the EBIC under-selects in the high dimensional setting.

Several authors have proposed methods which validate graph structure by quantification of the consistency of the models produced from data subsamples. The Stability Selection approach was proposed by Meinshausen and Buhlmann (13). Subsequently, the Stability Approach to Regularisation and Selection was proposed by Liu et al. (3). In StARS the highest regularisation parameter is selected such that the variability of the graph is less than a specified threshold. The assignment of the graph variability cut-off is typically fixed at 0.05 or 0.1. The stability methods require fitting of multiple models on data subsamples. Consequently, they underutilise the available data and present considerable computational demands. Nonetheless, these methods have demonstrated good performance in the high dimensional setting.

Rotational invariance selection was proposed by Lysen (7). Under this method, the input data matrix is concatenated with a row-permuted version. The permuted variables are thus known to be independent of the unpermuted variables. Variable selection is subsequently performed using LASSO. The least amount of regularisation is applied such that the permuted variables are not selected as predictors. This method is computationally efficient, as it spares inference of GLASSO with multiple regularisation values, however, it tends to select excessively sparse models in the high dimensional setting (16).

# Methodology

### Preliminaries

Let $X = \{X_{:,1}, \ldots X_{:,p}\}^T$ be a random vector of multivariate Gaussian distribution with covariance $\Sigma$. Let $X_{:,w}$ have zero mean and unit variance for all $w \in \{1, \ldots, p\}$. Let there be $n$ observations of $X$. Lauritzen demonstrated that a gaussian graphical model of $X$ may be recovered from the non-zero entries of its inverse covariance matrix(17). This property is a result of the relationship between entries of the inverse covariance matrix and the partial correlation coefficient, as $\rho_{kl} = \frac{-\Sigma^{-1}_{kl}}{\sqrt{\Sigma^{-1}_{kk}\Sigma^{-1}_{ll}}}$ (18). Consequently, independence of $V_k$ and $V_l$ is denoted by $\Sigma^{-1}_{kl} = 0$.

Let $G(V, E)$ be a graphical model generated from $\Sigma^{-1}$ such that each variable $X_{:,w}$, $w \in \{1, \ldots, p\}$, is represented by a vertex $V_w$ in $V$. Therefore, a perfect mapping exists from each variable of $X$ and $V$.

Let variables $V_k$ and $V_l$ be conditionally independent if no edge exists between $V_k$ and $V_l$ in $G$ $\{k, l\} \subset V$. Let variables $V_k$ and $V_l$ be conditionally dependent if they are adjacent in $G$. Therefore, a perfect mapping exists from each conditional dependency in $X$ and $E$.

Let $A$ be the binary adjacency matrix of $G$.

$$A_{kl} = \begin{cases} 1, & \Sigma^{-1}_{kl} \neq 0 \\ 0, & \Sigma^{-1}_{kl} = 0 \\ 0, & k = l \end{cases}$$

$A_{kl} = 0, \forall \{k, k \in V\}$, i.e. no self-loops exist.

Let $\Gamma_w$ be the set of nodes which are adjacent to $V_w$ in $G$. Therefore, $\Gamma_w = \{A_{:,w}, A_{:,w} = 1\}$.

The Local Markov property states that any variable is conditionally independent of all other variables, given the variables corresponding to its adjacent vertices (18).

$$X_{:,w} \perp\!\!\!\perp X_{:,V\setminus\{w,\Gamma_w\}} | X_{:,\Gamma_w}$$

We seek to test the faithfulness of $G$ to the local Markov property in the joint distribution of $X$.

$A$ is a binary simplification of $\Sigma^{-1}$, allowing the distinction of conditionally independent variables from conditionally dependent variables. $A_{ij}$ is invariant to $\Sigma^{-1}_{ij}$, given that $\Sigma^{-1}_{ij} \neq 0$. Nonetheless, $A$ contains sufficient information to recover all Markov blankets in $G$. Accordingly, we proceed to test the faithfulness of $G$ to the conditional dependence structure of $X$ using $A$. In accordance with the binary domain of $A$, we seek a measure in which each conditional dependency of $X_{:,\Gamma_w}$ is assigned uniform importance.

The binary adjacency matrix is a simplification of the model of $X$ as any two non-zero values of $\Sigma^{-1}$ are represented with 1. In the binary graph all active variables will adjacent to the response variable. Conversely, Inactive variables will be at a geodesic distance greater than one. The adjacency matrix therefore contains all information required to recover the Markov blanket of any variable in $G$.

## Purpose of Model

We propose to evaluate the performance of a model which estimates the values of variable $X_{:,w}$ given the subspace of $X$ corresponding to $\Gamma_w$ in $G$, for $w \in \{1, \dots, p\}$. $\Gamma_w$ corresponds to the *Markov Blanket* of $V_w$ (18–20). If the Local Markov property is true on $G$, then $X_{:,V\setminus\{w,\Gamma_w\}}$ contains as much information on $X_{:,w}$ as $X_{:,V\setminus\{w\}}$. . For the purposes of our algorithm, we seek to evaluate the dependence relationships between $X_{:,w}$ and $X_{:,\Gamma_w}$ for $w \in \{1, \dots, p\}$. The cardinality of the neighbourhood of $V_w$ varies according to graph topology and sparsity. Thus, we require a test of dependence between a univariate response vector $X_{:,w}$ and a multivariate vector of its active variables $X_{:,\Gamma_w}$.

## Graphical Neighbour Information Algorithm Pseudocode

We propose a novel model selection criterion, Graphical Neighbour Information. This method provides a goodness-of-fit score of a binary graph structure to a given dataset.

### *Input*

    data matrix: $X \in \mathbb{R}^{n \times p}$

    set $\mathcal{A}$ of graph adjacency matrices such that $\mathcal{A}_i = A \in \{1,0\}^{p \times p}$

### *Output*

    $\mathcal{A}_{optimal}$, the adjacency matrix in $\mathcal{A}$ which is best fitted to $X$.

### *Procedure*

1. Sample $m$ observations of $X$ with replacement, $X^{b_1}$. $X^{b_1} \in \mathbb{R}^{m \times p}$.
2. Sample $m$ observations of $X$ with replacement, $X^{b_2}$. $X^{b_2} \in \mathbb{R}^{m \times p}$.
3. $X^b \leftarrow |X^{b_1} - X^{b_2}|$. $X^b \in \mathbb{R}^{m \times p}$.
4. Standardise $X^b$.
5. FOR $\mathcal{A}_i \in \mathcal{A}$:
    i. FOR $w$ in $\{1, \dots, p\}$:
        i. Find $\Gamma_w$, the set of adjacent nodes to $V_w$ in $\mathcal{A}$.
        ii. Estimate each variable $X^b_{:,w}$ as the mean of the adjacent nodes in $\mathcal{A}_i$.
        iii. $\widehat{X^b_{:,w}} \leftarrow |\Gamma_w|^{-1} \Sigma\, X^b_{\Gamma_w}$
        iv. $if\ |\Gamma_w| = 0\ then\ \widehat{X^b_{:,w}} \leftarrow 0_n$
        v. END FOR
    ii. Compute $MSE_{model}$, the mean squared error of the estimate.
    iii. Compute $\mathbb{E}[MSE_{random}]$, the expected $MSE$ of a random permutation of the model.
    iv. $Graphical\ Neighbour\ Information_{\mathcal{A}_i,X} \leftarrow \mathbb{E}[MSE_{random}] - MSE_{model}$
    v. END FOR
6. Select $\mathcal{A}_{optimal}$ to maximise $Graphical\ Neighbour\ Information_{\mathcal{A}_{optimal},X}$

## Nearest Neighbour Regression

Considering this requirement, fixed-radius nearest neighbour regression is a natural fit. Under this method, similarity is measured in terms of geometric distance between the predictor variables of the observations.

Neighbourhood regression is a non-parametric method which allows for non-linear manifold modelling. It assumes that the probability distribution of new observations follows a locally constant function (21). Using a fixed radius threshold $h$ and the $\ell_1$ distance norm:

$$\widehat{X_{i,w}} = \begin{cases} \dfrac{\Sigma_n \mathbb{1}\{|X_{i,\backslash w} - X_{j,\backslash w}| \leq h\} X_{j,w}}{\Sigma_n \mathbb{1}\{|X_{i,\backslash w} - X_{j,\backslash w}| \leq h\}}, & if\ \Sigma_n \mathbb{1}\{|X_{i,\backslash w} - X_{j,\backslash w}| \leq h\} > 0 \\ 0, & if\ \Sigma_n \mathbb{1}\{|X_{i,\backslash w} - X_{j,\backslash w}| \leq h\} = 0 \end{cases}$$

Where $\mathbb{1}\{condition_a\}$ denotes the indicator function that $condition_a$ is true.

The properties of this approach have been extensively studied. There are several known events under which the model will perform poorly (22):

I. Insufficient $|\Sigma_n \mathbb{1}\{|X_{i,\backslash w} - X_{j,\backslash w}| \leq h\}|$. Predictions generated from a small number of neighbouring instances are susceptible to noise . Consequently, the model suffers high variance and $\mathbb{E}[X_{i,w} - \widehat{X_{i,w}}]$ is large.
II. Excessive $|\Sigma_n \mathbb{1}\{|X_{i,\backslash w} - X_{j,\backslash w}| \leq h\}|$. Predictions generated from many samples are biased towards the mean. Therefore, $\mathbb{E}[X_{i,w} - \widehat{X_{i,w}}] \to 0$.
III. $|\Sigma_n \mathbb{1}\{|X_{i,\backslash w} - X_{j,\backslash w}| \leq h\}| = 0$. No samples are available for prediction.

## Nearest Neighbour Regression on Graphs

In the graphical analogy of the fixed radius neighbourhood regression, the $|X_{i,\backslash w} - X_{j,\backslash w}|$ metric is substituted with $d_G(V_i, V_j)$, the geodesic distance between the vertices representing $X_{i,:}$ and $X_{j,:}$. If $h$ is set to 1 in a binary graph, then $A$ contains the neighbourhood of each node.

$$\therefore \widehat{X_{:,w}} := \frac{X_{:,w} A_{w,:}}{\Sigma A_{w,:}}$$

This estimate is undefined in the case that an observation is isolated in the graph, as in this case, $\Sigma A_{w,:} = 0$.

Graphical covariance modelling generates a graph in which vertices represent variables and observations are represented by vertex attributes. This representation differs from the graphical neighbour regression configuration where vertices represent observations and variables are represented by vertex attributes. We henceforth refer to these graph configurations as "variable-oriented" and "observation-oriented".

Nonetheless, Graphical covariance modelling has analogous weakness to those of local regression.

I. Sparse models underfit the joint probability distribution of $X$, attributing a large proportion of the variability of $X_{:,w}$ to random noise. Under a sparse model:
$$\widehat{X_{:,w}} := f(X_{:,\Gamma_w}) + \epsilon_w, |\Gamma_w| \gtrsim 0$$

II. Dense models overfit the joint probability distribution of $X$, attributing variability of $X_{:,w}$ to numerous apparent dependencies in $X_{:,\backslash w}$. In dense graphical models, many instances are included in $\Gamma_w$, such that:
$$\widehat{X_{:,w}} := f(X_{:,\Gamma_w}) + \epsilon_w, |\Gamma_w| \gg 0$$

III. Variables represented by isolated vertices are considered independent of all other variables. Therefore $f(X_{:,\backslash w})$ contains no information on $X_{:,w}$ and:
$$\widehat{X_{:,w}} := f(X_{:,\Gamma_w}) + \epsilon_w, |\Gamma_w| = 0$$
$$\therefore \widehat{X_{:,w}} = \epsilon_w$$

A critical distinction between variable-oriented graphs and observation-oriented graphs is found in the local similarity of the vertices. The observation-oriented graph estimates the distance between vertex attributes as a function of geodesic distance. Thus, adjacent vertices are expected to have similar attributes. In contrast, a variable-oriented graph could not operate on a similar assumption. Adjacent vertices in variable-oriented graphs represent dependent variables. Even in the case that the variables were standardised to equate variance, variables with negative covariance would be expected to have a higher distance between observations than independent variables.

Consider a standardised joint distribution. The expectation of the distance between observations of negatively covariant variables would exceed the expectation of distance between independent variables. Clearly this phenomenon would violate the assumption that distance between vertex attributes grows as a function of geodesic distance. In the case that the covariance matrix was constrained such that $\Sigma_{wz} \geq 0, \forall \{w, z\} \subset \{1, \dots p\}$, it may be expected that dependent variables will have a lower distance than independent variables. This is an unreasonable constraint to the task of graph inference, given that the true covariance matrix is generally unknown.

## Adapting Graphical Neighbour Regression to Variable-Oriented Graphs

In order to deploy a nearest neighbour model on a variable-oriented graphs, we must conform to the assumptions of the model. We require a transformation of $X$ which preserves its dependency structure whilst guaranteeing universally positive covariance. We approach this problem by sampling from the pairwise absolute differences between observations of $X$.

We generate $X^b$, a matrix of scaled absolute differences between sampled observations of $X$, as follows:

1. Set the number of samples, $m$
2. Generate $X^{b_1}$. Sample $m$ observations from $X$ with replacement.
3. Generate $X^{b_2}$. Sample $m$ observations from $X$ with replacement.
4. $X^b_{i,j} \leftarrow |X^{b_1}_{i,j} - X^{b_2}_{i,j}| \; \forall \; i \in \{1, \ldots, m\}, j \in \{1, \ldots, p\}$
5. $X^b_{i,j} \leftarrow \frac{X^b_{i,j} - \mu(X^b_{:,j})}{sd(X^b_{:,j})}$

Standardisation of each variable in $X^b$ ensures that it has zero mean and unit variance for each variable. $X^b$ bears some similarity to the pairwise distance matrix used by Szekely et al for the computation of distance correlation (23), with several notable differences.

1. Distance is measured under the $\ell_1$ norm. This facilitates rapid computation of the score for large matrices and demonstrates excellent performance. The Euclidean $\ell_2$ norm is used in the distance covariance score.
2. $X^b$ is not row-normalised.

## Properties of Graphical Neighbour Regression

Let $\mu_k = \mu_l$

Let $X_{:,w} = rX_{:,k} + \epsilon_w$, where

1. $\epsilon_w$ is fixed vector of errors which is independent of $X_{:,w}$ and $X_{:,k}$, such that $\mu\epsilon_w = 0$ and $\epsilon_W \perp \{X_{:,w}, X_{:,k}\}$.
2. $X_{:,k}$ is a univariate vector, such that $k \in \{1, \ldots, p\}, k \neq w$.

Let

$$\mathbb{E}[\varrho_{w,k}] := \mathbb{E}\left[(|X_{i,w} - X_{j,w}| - |X_{i,k} - X_{j,k}|)^2\right]$$

Therefore, $\varrho_{w,k}$ is non-negative. This satisfies the first requirement of our function.

$$\mathbb{E}[\varrho_{w,k}] = \mathbb{E}\left[|X_{i,w} - X_{j,w}|^2 + |X_{i,k} - X_{j,k}|^2 - 2|(X_{i,w} - X_{j,w})(X_{i,k} - X_{j,k})|\right]$$

$$= \mathbb{E}\left[(X_{i,w} - X_{j,w})^2\right] + \mathbb{E}\left[(X_{i,k} - X_{j,k})^2\right] - (2)\mathbb{E}[|(X_{i,w} - X_{j,w})(X_{i,k} - X_{j,k})|]$$

As $X_{:,k} \sim N(0,1)$

$$\therefore (X_{i,k} - X_{j,k}) \sim N(0,2)$$

$$\therefore (X_{i,k} - X_{j,k})^2 \sim \chi^2(2)$$

$$\mathbb{E}\left[(X_{i,k} - X_{j,k})^2\right] = \mathbb{E}[\chi^2(2)] = 2$$

Likewise, as $X_{:,w} \sim N(0,1)$

$$\mathbb{E}\left[(X_{i,w} - X_{j,w})^2\right] = 2$$

$$\therefore \mathbb{E}[\varrho_{w,k}] = \mathbb{E}\left[(X_{i,w} - X_{j,w})^2\right] + \mathbb{E}\left[(X_{i,k} - X_{j,k})^2\right] - 2 * \mathbb{E}[|(X_{i,w} - X_{j,w})(X_{i,k} - X_{j,k})|]$$

$$= 2 + 2 - 2 * \mathbb{E}\left[|\left((rX_{i,k} + \epsilon_{i,w}) - (rX_{j,k} + \epsilon_{j,w})\right)(X_{i,k} - X_{j,k})|\right]$$

$$= 4 - 2 * \mathbb{E}\left[|\left(r(X_{i,k} - X_{j,k}) + (\epsilon_{i,w} - \epsilon_{j,w})\right)(X_{i,k} - X_{j,k})|\right]$$

$$= 4 - 2 * \mathbb{E}\left[|r(X_{i,k} - X_{j,k})^2 + (\epsilon_{i,w} - \epsilon_{j,w})(X_{i,k} - X_{j,k})|\right]$$

Also

$$\sigma^2[(\epsilon_{i,w} - \epsilon_{j,w})(X_{i,k} - X_{j,k})] =$$

$$\mathbb{E}\left[(\epsilon_{i,w} - \epsilon_{j,w})^2(X_{i,k} - X_{j,k})^2\right] - (\mathbb{E}[(\epsilon_{i,w} - \epsilon_{j,w})(X_{i,k} - X_{j,k})])^2 =$$

$$\sigma^2[(\epsilon_{i,w} - \epsilon_{j,w})]\sigma^2[(X_{i,k} - X_{j,k})] + \sigma^2[(\epsilon_{i,w} - \epsilon_{j,w})](\mathbb{E}[(\epsilon_{i,w} - \epsilon_{j,w})])^2$$
$$+ \sigma^2[(X_{i,k} - X_{j,k})](\mathbb{E}[(\epsilon_{i,w} - \epsilon_{j,w})])^2 =$$

$$\sigma^2[(\epsilon_{i,w} - \epsilon_{j,w})]\sigma^2[(X_{i,k} - X_{j,k})] + \sigma^2[(\epsilon_{i,w} - \epsilon_{j,w})] * 0 + \sigma^2[(X_{i,k} - X_{j,k})] * 0 =$$

$$\sigma^2[(\epsilon_{i,w} - \epsilon_{j,w})]\sigma^2[(X_{i,k} - X_{j,k})] = 4\sigma^2_{\epsilon_w} * 4 = 16\sigma^2_{\epsilon_w}$$

$$\therefore (\epsilon_{i,w} - \epsilon_{j,w})(X_{i,k} - X_{j,k}) \sim N(0, 16\sigma_{\epsilon_w}^2)$$

$$\therefore r(X_{i,k} - X_{j,k})^2 + (\epsilon_{i,w} - \epsilon_{j.w})(X_{i,k} - X_{j,k}) \sim N\left(0, (2r + 16\sigma_{\epsilon_w}^2)\right)$$

$$\therefore \mathbb{E}\left[\left|r(X_{i,k} - X_{j,k})^2 + (\epsilon_{i,w} - \epsilon_{j.w})(X_{i,k} - X_{j,k})\right|\right] = (2r + 16\sigma_{\epsilon_w}^2)^2 \sqrt{\frac{2}{\pi}}$$

$$= 4\sqrt{\frac{2}{\pi}}(r + 8\sigma_{\epsilon_w}^2)^2$$

$$\therefore \mathbb{E}[\varrho_{w,k}] = 4 - 8\sqrt{\frac{2}{\pi}}(r + 8\sigma_{\epsilon_w}^2)^2$$

Assuming that $\sigma_{\epsilon_w}$ is small (with respect to $r$), the expectation of $\varrho_{w,k}$ is inversely associated with the $r^2$. In the case that $\sigma_{\epsilon_w}$ is non-negligible this function is quadratic, minimising with $r < 0$. Under these conditions, $\varrho_{w,k}$ fulfils our second requirement.

In the case of graphical model inference, we are interested in testing the dependence of each variable $V_w$ against its Markov blanket. As $|\Gamma_w| \in \{0, \ldots, p\}$ we require that the calculation supports the evaluation of nodes with multiple active variables.

Let

$$X_{:,w} = (r_1 X_{:,1} + r_2 X_{:,2} + \cdots r_{|Z|} X_{:,|Z|} + \epsilon_w) = \sum_{k \in Z} r_k X_{:,k} + \epsilon_w$$

Let

$$\mathbb{E}[\varrho_{w,k}] := \mathbb{E}\left[\left(|X_{i,w} - X_{j,w}| - \left||Z|^{-1}\sum_{k \in Z}(r_k X_{i,k} - r_k X_{j,k})\right|\right)^2\right]$$

$$= \mathbb{E}\left[(X_{i,w} - X_{j,w})^2\right] + \mathbb{E}\left[\left(|Z|^{-1}\sum_{k \in Z}(r_k X_{i,k} - r_k X_{j,k})\right)^2\right]$$

$$- (2)\mathbb{E}\left[\left|(X_{i,w} - X_{j,w})\left(|Z|^{-1}\sum_{k \in Z}(r_k X_{i,k} - r_k X_{j,k})\right)\right|\right]$$

$$|Z|^{-1}\sum_{k \in Z} r_k(X_{i,k} - X_{j,k}) \sim |Z|^{-1}\sum_{k \in Z} N(0,2) = \frac{N(0, 2|Z|)}{|Z|}$$

$$\mathbb{E}\left[\left(|Z|^{-1}\sum_{k \in Z} r_k(X_{i,k} - X_{j,k})\right)^2\right] = \frac{\mathbb{E}[\chi^2(2|Z|)]}{|Z|} = 2$$

Therefore

$$\mathbb{E}[\varrho_{w,k}] = 2 + 2 - 2$$
$$* \mathbb{E}\left[\left(\left(\sum_{k \in Z}(r_k \boldsymbol{X}_{i,k}) + \epsilon_{i,w}\right) - \left(\sum_{k \in Z}(r_k \boldsymbol{X}_{j,k}) + \epsilon_{j,w}\right)\right)\left(|Z|^{-1} \sum_{k \in Z} r_k(\boldsymbol{X}_{i,k} - \boldsymbol{X}_{j,k})\right)\right]$$

$$= 4 - 2\mathbb{E}\left[\left(\sum_{k \in Z} r_k(\boldsymbol{X}_{i,k} - \boldsymbol{X}_{j,k}) + (\epsilon_{i,w} - \epsilon_{j,w})\right)\left(|Z|^{-1} \sum_{k \in Z} r_k(\boldsymbol{X}_{i,k} - \boldsymbol{X}_{j,k})\right)\right]$$

$$= 4 - 2|Z|^{-1}\mathbb{E}\left[\left(\sum_{k \in Z} r_k(\boldsymbol{X}_{i,k} - \boldsymbol{X}_{j,k})\right)^2 + \left(\sum_{k \in Z} r_k(\boldsymbol{X}_{i,k} - \boldsymbol{X}_{j,k})\right)(\epsilon_{i,w} - \epsilon_{j,w})\right]$$

Also

$$\sigma^2\left[(\epsilon_{i,w} - \epsilon_{j,w})\left(\sum_{k \in Z} r_k(\boldsymbol{X}_{i,k} - \boldsymbol{X}_{j,k})\right)\right] =$$

$$\mathbb{E}\left[(\epsilon_{i,w} - \epsilon_{j,w})^2 \left(\sum_{k \in Z} r_k(\boldsymbol{X}_{i,k} - \boldsymbol{X}_{j,k})\right)^2\right] - \left(\mathbb{E}\left[(\epsilon_{i,w} - \epsilon_{j,w})\left(\sum_{k \in Z} r_k(\boldsymbol{X}_{i,k} - \boldsymbol{X}_{j,k})\right)\right]\right)^2 =$$

$$\sigma^2[(\epsilon_{i,w} - \epsilon_{j,w})]\sigma^2\left[\left(\sum_{k \in Z} r_k(\boldsymbol{X}_{i,k} - \boldsymbol{X}_{j,k})\right)\right] + \sigma^2[(\epsilon_{i,w} - \epsilon_{j,w})](\mathbb{E}[(\epsilon_{i,w} - \epsilon_{j,w})])^2$$
$$+ \sigma^2\left[\left(\sum_{k \in Z} r_k(\boldsymbol{X}_{i,k} - \boldsymbol{X}_{j,k})\right)\right](\mathbb{E}[(\epsilon_{i,w} - \epsilon_{j,w})])^2 =$$

$$\sigma^2[(\epsilon_{i,w} - \epsilon_{j,w})]\sigma^2\left[\left(\sum_{k \in Z} r_k(\boldsymbol{X}_{i,k} - \boldsymbol{X}_{j,k})\right)\right] = 4\sigma_{\epsilon_w}^2 * 4|Z| = 16|Z|\sigma_{\epsilon_w}^2$$

$$\sigma^2[(\epsilon_{i,w} - \epsilon_{j,w})]\sigma^2\left[\sum_{k \in Z} r_k(\boldsymbol{X}_{i,k} - \boldsymbol{X}_{j,k})\right] + \sigma^2[(\epsilon_{i,w} - \epsilon_{j,w})] * 0 + \sigma^2\left[\sum_{k \in Z} r_k(\boldsymbol{X}_{i,k} - \boldsymbol{X}_{j,k})\right] * 0 =$$

$$\therefore (\epsilon_{i,w} - \epsilon_{j,w})\left(\sum_{k \in Z} r_k(\boldsymbol{X}_{i,k} - \boldsymbol{X}_{j,k})\right) \sim N(0, 16|Z|\sigma_{\epsilon_w}^2)$$

$$\therefore \left(\left(\sum_{k \in Z} r_k(\boldsymbol{X}_{i,k} - \boldsymbol{X}_{j,k})\right)^2 + \left(\sum_{k \in Z} r_k(\boldsymbol{X}_{i,k} - \boldsymbol{X}_{j,k})\right)(\epsilon_{i,w} - \epsilon_{j,w})\right) \sim N\left(0, (2|Z|\Sigma r_k + 16|Z|\sigma_{\epsilon_w}^2)\right)$$

The expectation of the absolute value may be calculated from the folded normal $N\left(0, \left(2|Z|\Sigma r_k + 16|Z|\sigma_{\epsilon_w}^2\right)\right)$.

$$\mathbb{E}\left[\left|\left(\sum_{k\in Z} r_k(X_{i,k} - X_{j,k})\right)^2 + \left(\sum_{k\in Z} r_k(X_{i,k} - X_{j,k})\right)(\epsilon_{i,w} - \epsilon_{j,w})\right|\right] = \left(2|Z|\Sigma r_k + 16|Z|\sigma_{\epsilon_w}^2\right)^2 \sqrt{\frac{2}{\pi}}$$

$$= 4 - 2|Z|^{-1}\left(2|Z|\sum_{k\in Z} r_k + 16|Z|\sigma_{\epsilon_w}^2\right)^2 \sqrt{\frac{2}{\pi}}$$

$$\therefore \mathbb{E}[\varrho_{w,k}] = 4 - 8\sqrt{\frac{2}{\pi}}|Z|\left(\sum_{k\in Z} r_k + 8\sigma_{\epsilon_w}^2\right)^2$$

Therefore $\mathbb{E}[\varrho_{w,k}]$ is maximised when $\Sigma r_k = -8\sigma_{\epsilon_w}^2$. If the variance of the error is small with respect to $\Sigma r_k$ then $\mathbb{E}[\varrho_{w,k}]$ decreases as $|\Sigma r_k|$ increases. In this condition, $\mathbb{E}[\varrho_{w,k}]$ varies inversely with the expected linear fit of a linear model of $X_{:,w}$ on $X_{:,Z}$, in which all coefficients are fixed at $|Z|^{-1}$. Therefore, we propose that $\mathbb{E}[\varrho_{w,k}]$ is an appropriate model of goodness-of-fit of the binary graph $G$ on $X$.

Let $|Z| = |\Gamma_w|$.

$$\mathbb{E}[\varrho_{w,Z} - \varrho_{w,\Gamma_w}] = \left(4 - 8\sqrt{\frac{2}{\pi}}|Z|\left(\sum_{k\in Z} r_k + 8\sigma_{\epsilon_w}^2\right)^2\right) - \left(4 - 8\sqrt{\frac{2}{\pi}}|\Gamma_w|\left(\sum_{k\in \Gamma_w} r_k + 8\sigma_{\epsilon_w}^2\right)^2\right)$$

$$= -8\sqrt{\frac{2}{\pi}}|\Gamma_w|\left(\left(\sum_{k\in Z} r_k + 8\sigma_{\epsilon_w}^2\right)^2\right) - \left(\left(\sum_{k\in \Gamma_w} r_k + 8\sigma_{\epsilon_w}^2\right)^2\right)$$

$$= -8\sqrt{\frac{2}{\pi}}|\Gamma_w|\left(\left(\sum_{k\in Z} r_k\right)^2 + 16\sum_{k\in Z} r_k \sigma_{\epsilon_w}^2 + 64\sigma_{\epsilon_w}^4\right) - \left(\left(\sum_{k\in \Gamma_w} r_k\right)^2 + 16\sum_{k\in \Gamma_w} r_k \sigma_{\epsilon_w}^2 + 64\sigma_{\epsilon_w}^4\right)$$

$$= -8\sqrt{\frac{2}{\pi}}|\Gamma_w|\left(\left(\sum_{k\in Z} r_k\right)^2 + 16\sum_{k\in Z} r_k \sigma_{\epsilon_w}^2\right) - \left(\left(\sum_{k\in \Gamma_w} r_k\right)^2 + 16\sum_{k\in \Gamma_w} r_k \sigma_{\epsilon_w}^2\right)$$

$$= 8\sqrt{\frac{2}{\pi}}|\Gamma_w|\left(\left(\sum_{k\in \Gamma_w} r_k\right)^2 - \left(\sum_{k\in Z} r_k\right)^2\right) + 16\left(\sum_{k\in \Gamma_w} r_k \sigma_{\epsilon_w}^2 - \sum_{k\in Z} r_k \sigma_{\epsilon_w}^2\right)$$

$$= 8\sqrt{\frac{2}{\pi}}|\Gamma_w|\left(\left(\left(\sum_{k\in\Gamma_w\cap Z}r_k\right)^2 + \left(\sum_{k\in\Gamma_w\setminus Z}r_k\right)^2 + 2\left(\sum_{k\in\Gamma_w\cap Z}r_k\right)\left(\sum_{k\in\Gamma_w\setminus Z}r_k\right)\right)\right.$$

$$\left.-\left(\left(\sum_{k\in\Gamma_w\cap Z}r_k\right)^2 + \left(\sum_{k\in Z\setminus\Gamma_w}r_k\right)^2 + 2\left(\sum_{k\in\Gamma_w\cap Z}r_k\right)\left(\sum_{k\in Z\setminus\Gamma_w}r_k\right)\right)\right)$$

$$+ 16\left(\sum_{k\in\Gamma_w}r_k\sigma^2_{\epsilon_w} - \sum_{k\in Z}r_k\sigma^2_{\epsilon_w}\right)$$

$$=\left(\left(\left(\sum_{k\in\Gamma_w\setminus Z}r_k\right)^2 + 2\left(\sum_{k\in\Gamma_w\cap Z}r_k\right)\left(\sum_{k\in\Gamma_w\setminus Z}r_k\right)\right)\right.$$

$$\left.-\left(+\left(\sum_{k\in Z\setminus\Gamma_w}r_k\right)^2 + 2\left(\sum_{k\in\Gamma_w\cap Z}r_k\right)\left(\sum_{k\in Z\setminus\Gamma_w}r_k\right)\right)\right)$$

$$+ 16\left(\sum_{k\in\Gamma_w\setminus Z}r_k\sigma^2_{\epsilon_w} - \sum_{k\in Z\setminus\Gamma_w}r_k\sigma^2_{\epsilon_w}\right)$$

$$=\left(\sum_{k\in\Gamma_w\setminus Z}r_k\right)^2 - \left(\sum_{k\in Z\setminus\Gamma_w}r_k\right)^2 + 2\left(\sum_{k\in\Gamma_w\cap Z}r_k\right)\left(\sum_{k\in\Gamma_w\setminus Z}r_k - \sum_{k\in Z\setminus\Gamma_w}r_k\right)$$

$$+ 16\sigma^2_{\epsilon_w}\left(\sum_{k\in\Gamma_w\setminus Z}r_k - \sum_{k\in Z\setminus\Gamma_w}r_k\right)$$

$$=\left(\sum_{k\in\Gamma_w\setminus Z}r_k + \sum_{k\in Z\setminus\Gamma_w}r_k\right)\left(\sum_{k\in\Gamma_w\setminus Z}r_k - \sum_{k\in Z\setminus\Gamma_w}r_k\right) + 2\left(\sum_{k\in\Gamma_w\cap Z}r_k\right)\left(\sum_{k\in\Gamma_w\setminus Z}r_k - \sum_{k\in Z\setminus\Gamma_w}r_k\right)$$

$$+ 16\sigma^2_{\epsilon_w}\left(\sum_{k\in\Gamma_w\setminus Z}r_k - \sum_{k\in Z\setminus\Gamma_w}r_k\right)$$

$$=\left(\sum_{k\in\Gamma_w\setminus Z}r_k + \sum_{k\in Z\setminus\Gamma_w}r_k + 2\sum_{k\in\Gamma_w\cap Z}r_k + 16\sigma^2_{\epsilon_w}\right)\left(\sum_{k\in\Gamma_w\setminus Z}r_k - \sum_{k\in Z\setminus\Gamma_w}r_k\right)$$

$$=\left(\sum_{k\in Z}r_k + \sum_{k\in\Gamma_w}r_k + 16\sigma^2_{\epsilon_w}\right)\left(\sum_{k\in\Gamma_w\setminus Z}r_k - \sum_{k\in Z\setminus\Gamma_w}r_k\right)$$

$$if \left( \sum_{k \in \Gamma_w \setminus Z} r_k - \sum_{k \in Z \setminus \Gamma_w} r_k \right) \to 0$$

$$then \left( \sum_{k \in Z} r_k + \sum_{k \in \Gamma_w} r_k + 16\sigma_{\epsilon_w}^2 \right) \left( \sum_{k \in \Gamma_w \setminus Z} r_k - \sum_{k \in Z \setminus \Gamma_w} r_k \right) \to 0$$

Therefore $\mathbb{E}[\varrho_{w,Z} - \varrho_{w,\Gamma_w}]$ decreases as $Z \to \Gamma_w$. Thus, we propose that $\varrho_{w,Z} - \varrho_{w,\Gamma_w}$ provides a score of the fit of the model $\boldsymbol{X}_{:,\boldsymbol{w}} = f(Z)$. This score requires the assumption $|Z| = |\Gamma_w|$. The evaluation of this assumption is not tractable in general, as $|\Gamma_w|$ is unknown. Therefore, we compare the fit of estimated models to the fit of random models with equal complexity.

We exploit this phenomenon to appraise the quality of a given graphical model of a high dimensional dataset.

## Graphical Neighbour Information Criterion

We present the Graphical Neighbour Information Criterion.

Let $\widehat{\Gamma_w}$ be an estimate of set of active variables for $X_{:,w}$.

Let $\mathcal{H}$ be the hypothesis that $G$ encodes a graph which is faithful to $X$ with respect to the local Markov principle. $\mathcal{H}$ may be decomposed into $p$ sub-hypotheses, each pertaining to the local Markov property faithfulness of a specific variable. $\mathcal{H}$ is true if all sub-hypotheses are true. Let $\mathcal{H}_w$ be the sub-hypothesis relating to $X_{:,w}$.

$$\therefore \mathcal{H}_w := V_w \perp \{V_w \setminus \widehat{\Gamma_w}\} | \widehat{\Gamma_w}, \forall\, w \in \{1, \ldots, p\}$$

We proceed to test this hypothesis by comparing $G$ with randomly permuted models of equivalent complexity. We compare $\varrho_{w,\widehat{\Gamma_w}}$ to $\varrho_{w,Z}$, where $Z$ is a random sample of $V_{\setminus w}$, such that $Z \subset \{V \setminus w\}, |Z| \approx |\widehat{\Gamma_w}|$. Under the sparsity assumption, $|\Gamma_j| \ll (p-1)$. Therefore:

$$\mathbb{E}[|Z \cap \Gamma_j|] = |\Gamma_j|^{-1}|Z|^{-1} \approx 0$$

Therefore, we expect that $Z$ mostly contains vertices of which $X_{:,w}$ is conditionally independent. We expect that a graphical neighbour regression model trained $X_{:,\Gamma_w}$ should outperform an equivalent model trained on $X_{:,Z}$. Likewise, if $\widehat{\Gamma_w} = \Gamma_w$, then a model trained on $X_{:,\widehat{\Gamma_w}}$ should outperform a model trained on $X_{:,Z}$.

We reframe $\mathcal{H}_w$ as follows.

$$\mathcal{H}_w := \varrho_{w,\widehat{\Gamma_w}} < \varrho_{w,Z}$$

Let $GNI_{w,\widehat{\Gamma_w}}$ be the Graphical Neighbour Information of $X_{:,\widehat{\Gamma_w}}$ on $X_{:,w}$.

$$GNI_{w,\widehat{\Gamma_w}} := \varrho_{w,Z} - \varrho_{w,\widehat{\Gamma_w}}$$

We propose that

$$\mathbb{P}[\mathcal{H}_w] \propto GNI_{w,\widehat{\Gamma_w}}$$

Assuming that $V_k \perp V_l\, \forall\, k \in \widehat{\Gamma}_w$, then $\varrho_{w,\widehat{\Gamma_w}} - \varrho_{w,Z}$ is a function of the difference in fit of models on $\widehat{\Gamma}_w$ and $Z$. If $\varrho_{w,\widehat{\Gamma_w}} \approx \varrho_{w,Z}$ then $\widehat{\Gamma}_w$ carries a similar amount of information on $V_w$ as that which would be expected of a random model. Therefore, $\mathcal{H}_w$ cannot be rejected.

We deploy a graphical nearest neighbour model in which the value of $X_{i,w}^b$ is estimated by the mean of the corresponding observations in $X_{i,\Gamma_w}^b$ as follows:

$$\widehat{X_{i,j}^b} = \begin{cases} \dfrac{\sum X_{i,\Gamma_j}^b}{|\Gamma_j|}, & \text{if } |\Gamma_j| \neq 0 \\ 0, & \text{if } |\Gamma_j| = 0 \end{cases}$$

In the case that $V_w$ is isolated in $\hat{G}$, then $|\Gamma_w|$ and this value will consequently be undefined. We estimate all isolated vertices with $\mu(X_{:,\setminus w}^b)$, which is 0 due to the standardisation of $X^b$.

Complex models may tend to overfit data. Considering this phenomenon, we seek to appraise a models' information at a given level of complexity. The Bayesian and Akaike Information criteria explicitly assign penalties which correspond to the number of model parameters. We compare each model to the performance of random models with equivalent complexity. We define the adjusted model error as the difference in error between the query model and the expected error of random models with the same quantity and distribution of parameters. We generate random models by permuting the adjacency matrix and thereby randomly assigning the dependencies of $X$.

## Computation of the Graphical Neighbour Information

Let $\hat{G}$ be a model of $X$. Let $\hat{A}$ be the binary adjacency matrix of $\hat{G}$.

Let $\widehat{X^b_{i,:}}$ be the model estimate of $X_{i,:}$. We define the error of the model as the mean squared error. The vector of errors for the $i$th observation is defined as follows:

$$MSE_{model} = \frac{\left(\widehat{X^b} - X^b\right)^{\circ 2}}{mp}$$

Where "∘ 2" indicates the Hadamard index 2.

Let $d$ be a vector such that:

$$d_i \leftarrow f(x) = \begin{cases} |\Gamma_i|^{-1}, & if\ |\Gamma_i| > 0 \\ 1, & if\ |\Gamma_i| = 0 \end{cases}$$

Then

$$\widehat{X^b} = dX^b \hat{A}$$

We define an equivalent random model estimate of $X_{i,:}$ as a permutation of $\widehat{X^b_{i,:}}$.

Let $q$ be a random ordering of $\{1, \ldots, p\}$. Therefore, the mean squared error of the $q$-permuted model is:

$$MSE_{random^q_{i,j}} = \sum_{j=1}^{p} \frac{(\widehat{X^b_{i,q_j}} - X^b_{i,j})^2}{p}$$

Calculation of the true error of multiple permutations of the model is not necessary, as we require only the expected value of the set of all models. This may be calculated though the linearity of expectations.

$$\mathbb{E}\left[MSE_{random^q_{i,j}}\right] = \mathbb{E}\left[\sum_{j=1}^{p} \frac{(\widehat{X^b_{i,q_j}} - X^b_{i,j})^2}{p}\right]$$

$$= \mathbb{E}\left[\left(\widehat{X^b_{i,:}} - X^b_{i,:}\right)^{\circ 2}\right]$$

$$= \mathbb{E}\left[\widehat{X^b_{i,:}}^{\circ 2} + X^b_{i,:}{}^{\circ 2} - 2\widehat{X^b_{i,:}} X^b_{i,:}\right]$$

$$= \mathbb{E}\left[\widehat{X^b_{i,:}}^{\circ 2}\right] + \mathbb{E}\left[X^b_{i,:}{}^{\circ 2}\right] - 2\mathbb{E}\left[\widehat{X^b_{i,:}}\right]\mathbb{E}[X^b_{i,:}]$$

As the expectation of a vector is equivalent regardless of its permutation, we can now compute expected error of the permuted models in closed form, without calculating the individual models.

$$\mathbb{E}\left[MSE_{random^q_{:,j}}\right] = \mathbb{E}\left[\widehat{X^b_{i,:}}^2\right] + \mathbb{E}\left[X^b_{i,:}{}^2\right] - 2\mathbb{E}\left[\widehat{X^b_{i,:}}\right]\mathbb{E}[X^b_{i,:}]$$

$$= \mathbb{E}\left[\widehat{X^b_{i,:}}^2\right] + \mathbb{E}\left[X^b_{i,:}{}^2\right] - 2 * \mathbb{E}\left[\widehat{X^b_{i,:}}\right]\mathbb{E}[X^b_{i,:}]$$

And

$$\mathbb{E}[MSE_{random}] = \frac{\widehat{X^b}^{\circ 2}}{mp} + \frac{X^{b \circ 2}}{mp} - 2 \sum_{j=1}^{p} \frac{\widehat{X^b_{:,j}}}{p} \sum_{j=1}^{p} \frac{X^b_{:,j}}{p}$$

$$= \frac{\widehat{X^b}^{\circ 2} + X^{b \circ 2}}{mp} - \frac{2}{p^2} \sum_{j=1}^{p} \widehat{X^b_{:,J}} \sum_{j=1}^{p} X^b_{:,j}$$

$$\therefore \mathbb{E}\left[MSE_{random_{:,j}}\right] - MSE_{model} = \left(\frac{\widehat{X^b}^{\circ 2} + X^{b \circ 2}}{mp} - \frac{2}{p^2} \sum_{j=1}^{p} \widehat{X^b_{:,J}} \sum_{j=1}^{p} X^b_{:,j}\right) - \frac{\left(\widehat{X^b} - X^b\right)^{\circ 2}}{mp}$$

$$=$$

$$= \frac{\left(\widehat{X^b}^{\circ 2} + X^{b \circ 2}\right) - \left(\widehat{X^b}^{\circ 2} + X^{b \circ 2} - 2\widehat{X^b} \circ X^b\right)}{mp} - \frac{2}{p^2} \sum_{j=1}^{p} \widehat{X^b_{:,J}} \sum_{j=1}^{p} X^b_{:,j}$$

$$= 2\left(\frac{\widehat{X^b} \circ X^b}{mp} - p^{-2} \sum_{j=1}^{p} \widehat{X^b_{:,J}} \sum_{j=1}^{p} X^b_{:,j}\right)$$

This operation is therefore highly efficient, requiring two Hadamard products. The operation scales $O(p)$ under constant $n$. Assuming constant $p$ the operation scales in $m$, which is selected as a function of $n^2$.

# Simulations

We generated compared the performance of the Graphical Neighbourhood Information model selector to the current state of the art, using the implementations provided by the *huge* package.

## Data Generation Parameters

1. Data was generated from a sparse joint distribution using the "huge.generator" function.
2. Each dataset was generated with 50 observations.
3. 10 Datasets were generated with dimensionality of 50, 200 and 400. Therefore the $n:p$ ratios were 1, 0.25 and 0.125 respectively.
4. For each dimensionality, 5 graphs were generated according to the Erdos-Renyi random model with sparsity set at the package default of $\frac{p}{3}$. An additional 5 graphs were generated according to the hub, with hub count set at the package default of $\left\lceil \frac{p}{20} \right\rceil$.
5. For each dataset, a set $\mathcal{A}$ of candidate adjacency matrices was estimated, using the GLASSO algorithm with 30 regularisation parameters.
6. Model selection criteria were deployed to select the optimal graph. Rotation Invariance Criterion, EBIC, StARS and the Graphical Neighbourhood Information Criterion were deployed. For each of Rotation Invariance Criterion, EBIC and StARS, the implementations available in the package $huge$ were used.
7. The adjacency matrix of the selected graphs were compared to the true graph structure.
8. The model with the highest true F1 score was selected as an "oracle" comparison, indicating the hypothetical performance of GLASSO with optimal regularisation.

## Rotation Invariance Criterion

This method estimates the optimal regularisation parameter independently, taking only input data matrix. The optimal lambda was estimated for each given dataset and GLASSO was with a single parameter.

## StARS

This method requires GLASSO to be run on multiple subsamples of the data. The default number of subsamples, 25, was used. The graph with the highest variability less than a specific $\beta$ cut-off is selected. As per the package recommendations (24), the default value of 0.1 was used.

## EBIC

The Extended Bayesian Information Criterion has a single hyperparameter, $\gamma$, which controls the additional regularisation above the typical BIC. The authors (12) recommend a default regularisation of 0.5. In the original paper the values $\{0, 0.5, 1\}$ were used. Accordingly, we analyse the performance of EBIC with each of these parameters.

# Results

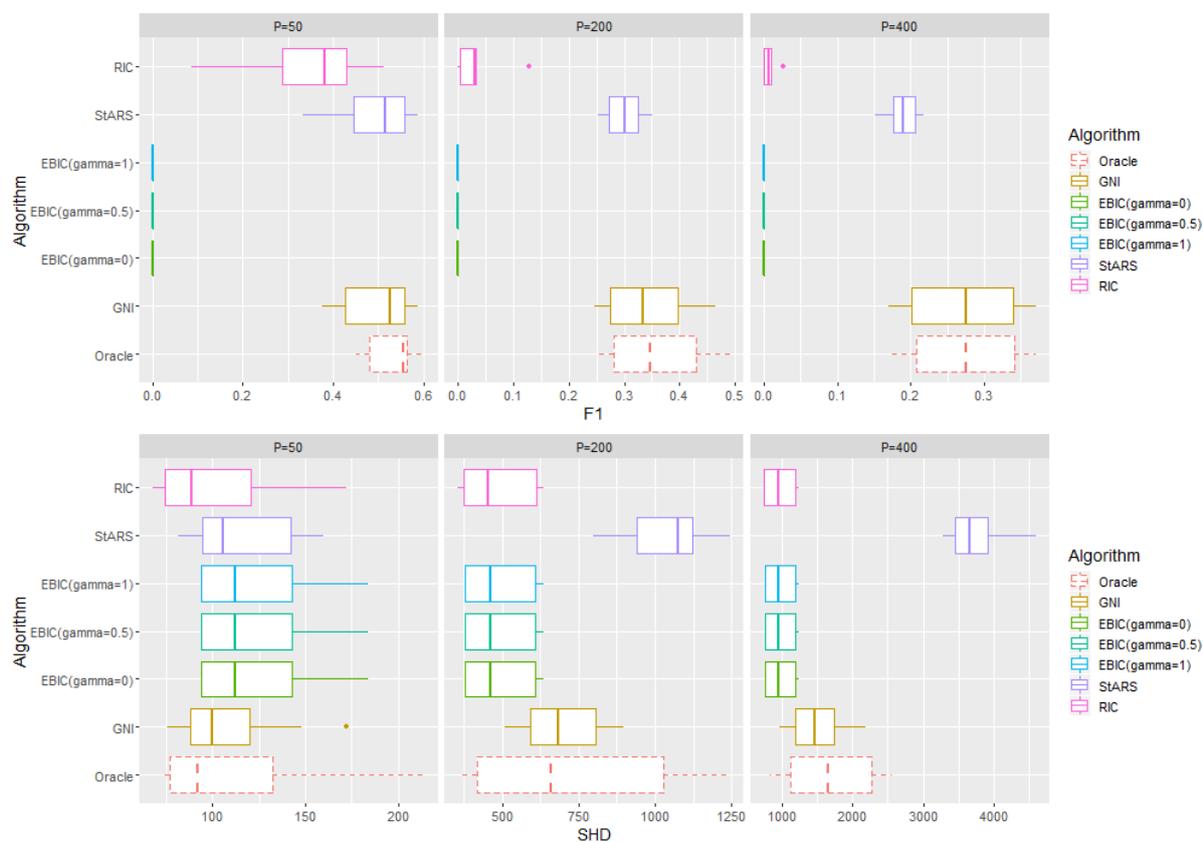

*Figure 1: Benchmark comparison of graphical model selection methods. Datasets were generated from multivariate distributions with 50 observations each. 10 datasets were generated with each of 50, 200 and 400 variables. 5 random graphs and 5 hub graphs of each dimensionality were generated. A candidate set of adjacency matrices was inferred using GLASSO with several regularisation levels. Selected models were compared to the true adjacency matrix of the generating distribution. Performance metrics are shown for a set of 5 random graphs and 5 hub structure graphs. The "oracle graph" was selected as the graph with the highest F1 score from the candidate set. Above: F1 score distribution of selected models. Below: Structural Hamming Distance of selected models.*

| Graph | P | Oracle | GNI | EBIC | StARS | RIC |
|---|---|---|---|---|---|---|
| random | 50 | 0.501354 | 0.440631 | 0.000000 | 0.445112 | 0.263431 |
| random | 200 | 0.277718 | 0.274204 | 0.000000 | 0.271423 | 0.008411 |
| random | 400 | 0.199967 | 0.194828 | 0.000000 | 0.175400 | 0.003333 |
| hub | 50 | 0.556827 | 0.550979 | 0.000000 | 0.545285 | 0.434960 |
| hub | 200 | 0.432053 | 0.407389 | 0.000000 | 0.326316 | 0.050361 |
| hub | 400 | 0.346288 | 0.343102 | 0.000000 | 0.203415 | 0.010433 |

*Figure 2: Mean F1 performance of each model. Mean values of 5 graphs are shown.*

| Graph  | P   | Oracle | GNI    | EBIC   | StARS  | RIC    |
|--------|-----|--------|--------|--------|--------|--------|
| random | 50  | 0.5570 | 0.6699 | 0.0000 | 0.6081 | 0.8732 |
| random | 200 | 0.2573 | 0.2978 | 0.0000 | 0.2225 | 0.3600 |
| random | 400 | 0.1706 | 0.2093 | 0.0000 | 0.1200 | 0.6000 |
| hub    | 50  | 0.6054 | 0.5434 | 0.0000 | 0.4997 | 0.8121 |
| hub    | 200 | 0.4550 | 0.3383 | 0.0000 | 0.2237 | 0.9857 |
| hub    | 400 | 0.3361 | 0.3049 | 0.0000 | 0.1217 | 0.8000 |

*Figure 3: Mean precision performance of each model*

| Graph  | P   | Oracle   | GNI      | EBIC     | StARS    | RIC      |
|--------|-----|----------|----------|----------|----------|----------|
| random | 50  | 0.473695 | 0.341531 | 0.000000 | 0.371167 | 0.160597 |
| random | 200 | 0.313886 | 0.254500 | 0.000000 | 0.348483 | 0.004268 |
| random | 400 | 0.243924 | 0.190523 | 0.000000 | 0.328219 | 0.001671 |
| hub    | 50  | 0.519149 | 0.570213 | 0.000000 | 0.604255 | 0.306383 |
| hub    | 200 | 0.424211 | 0.513684 | 0.000000 | 0.614737 | 0.026316 |
| hub    | 400 | 0.369474 | 0.396842 | 0.000000 | 0.625789 | 0.005263 |

*Figure 4: Mean recall performance of each model*

| Graph  | P   | Oracle | GNI    | EBIC   | StARS  | RIC    |
|--------|-----|--------|--------|--------|--------|--------|
| random | 50  | 146.8  | 131.6  | 154.4  | 138.0  | 134.4  |
| random | 200 | 1001.6 | 810.4  | 602.4  | 1126.0 | 600.8  |
| random | 400 | 2332.0 | 1871.6 | 1199.2 | 3703.6 | 1197.2 |
| hub    | 50  | 77.2   | 86.8   | 94.0   | 94.8   | 73.2   |
| hub    | 200 | 428.8  | 569.2  | 380.0  | 971.2  | 370.4  |
| hub    | 400 | 1066.4 | 1158.0 | 760.0  | 3751.6 | 756.0  |

*Figure 5: Mean Structural Hamming Distance of each model*

## Random graphs

For each dataset, the oracle graph was selected as the graph with the highest true F1 value. It functions as a measure of the performance of a hypothetical perfect selector. The oracle graph was moderately accurate in the low dimensional (precision = 0.557, recall = 0.474, F1 = 0.501, SHD = 146.800) and moderate dimensional simulations (precision = 0.257, recall = 0.314, F1 = 0.278, SHD = 1001.600). It was largely incorrect in the high dimensional setting (precision = 0.171, recall = 0.244, F1 = 0.200, SHD = 2332.000). RIC demonstrated adequate performance in the low dimensional setting on random graphs (precision = 9e-01, recall = 2e-01, F1 = 3e-01, SHD = 1e+02), which degraded in moderate dimensional (precision = 4e-01, recall = 4e-03, F1 = 8e-03, SHD = 6e+02) and high dimensional simulations (precision = 6e-01, recall = 2e-03, F1 = 3e-03, SHD = 1e+03) . StARS performed well in all simulations, achieving high performance in low dimensional graphs (precision = 0.608, recall = 0.371, F1 = 0.445, SHD = 138.000), moderate dimensional graphs (precision = 0.222, recall = 0.348, F1 = 0.271, SHD = 1126.000) and high dimensional graphs (precision = 0.120, recall = 0.328, F1 = 0.175, SHD = 3703.600). EBIC selected fully disconnected graphs in all simulations, with gamma settings at 0, 0.5 and 1. The Graphical Neighbour Information criterion selected equivalent graphs to the oracle solution in the low dimensional (precision = 0.670, recall = 0.342, F1 = 0.441, SHD = 131.600) and moderate dimensional settings (precision = 0.298, recall = 0.254, F1 = 0.274, SHD = 810.400). Furthermore, the Graphical Neighbour Information criterion matched the performance of the oracle in the high dimensional graphs (precision = 0.209, recall = 0.191, F1 = 0.195, SHD = 1871.600). The models selected by GNI in the high dimensional simulations did not have significantly higher F1 scores that those of StARS in random graphs (t = 1.474, 95% CI = [-0.01175, 0.05060], p.value = 0.18402), however, selections significantly outperformed StARS-selections in terms of Structural Hamming Distance (t = -9.796, 95% CI = [-2270,-1394], p.value = 1.7068e-05). GNI achieved oracle performance in terms of F1 (t = -0.3431, 95% CI = [-0.03971, 0.02943], p.value = 0.74042) and sub-oracle performance in terms of Structural Hamming Distance (t = -3.165, 95% CI = [-797.7,-123.1], p.value = 0.013843).

### Hub graphs

The oracle solution for hub graphs was moderately lower than that of random graphs in each of the low dimensional (precision = 0.605, recall = 0.519, F1 = 0.557, SHD = 77.200), moderate dimensional (precision = 0.455, recall = 0.424, F1 = 0.432, SHD = 428.800) and high dimensional settings (precision = 0.336, recall = 0.369, F1 = 0.346, SHD = 1066.400). RIC demonstrated adequate performance in the low dimensional setting on graphs (precision = 8e-01, recall = 3e-01, F1 = 4e-01, SHD = 7e+01), which again degraded in the moderate dimensional(precision = 1e+00, recall = 3e-02, F1 = 5e-02, SHD = 4e+02) and high dimensional graphs (precision = 8e-01, recall = 5e-03, F1 = 1e-02, SHD = 8e+02). StARS performed well in the low dimensional (precision = 0.500, recall = 0.604, F1 = 0.545, SHD = 94.800) and moderate dimensional graphs (precision = 0.224, recall = 0.615, F1 = 0.326, SHD = 971.200). However, performance degraded in the high dimensional setting (precision = 0.122, recall = 0.626, F1 = 0.203, SHD = 3751.600). EBIC selected fully disconnected graphs in all simulations, with gamma settings at 0, 0.5 and 1. The Graphical Neighbour Information criterion matched the oracle solution in low dimensional (precision = 0.543, recall = 0.570, F1 = 0.551, SHD = 86.800) and moderate dimensional settings (precision = 0.338, recall = 0.514, F1 = 0.407, SHD = 569.200) and high dimensional hub graphs (precision = 0.305, recall = 0.397, F1 = 0.343, SHD = 1158.000). High dimensional hub models selected by GNI demonstrated higher F1 score performance (t = 14.03, 95% CI = [0.1164,0.1629], p.value = 1.2013e-06) and lower Structural Hamming Distance loss (t = -11.5, 95% CI = [-3195,-1993], p.value = 0.00017303) than those selected by StARS. GNI achieved oracle performance in terms of F1 (t = -0.2738, 95% CI = [-0.03004, 0.02367], p.value = 0.79122) and sub-oracle performance in terms of Structural Hamming Distance (t = -3.165, 95% CI = [-797.7,-123.1], p.value = 0.013843) in high dimensional hub graphs.

We cannot exclude the possibility that superior results may have been demonstrated by the EBIC with different regularisation parameters. However, to the knowledge of the authors, specific guidance on regularisation selection outside of the [0,1] range is unavailable.

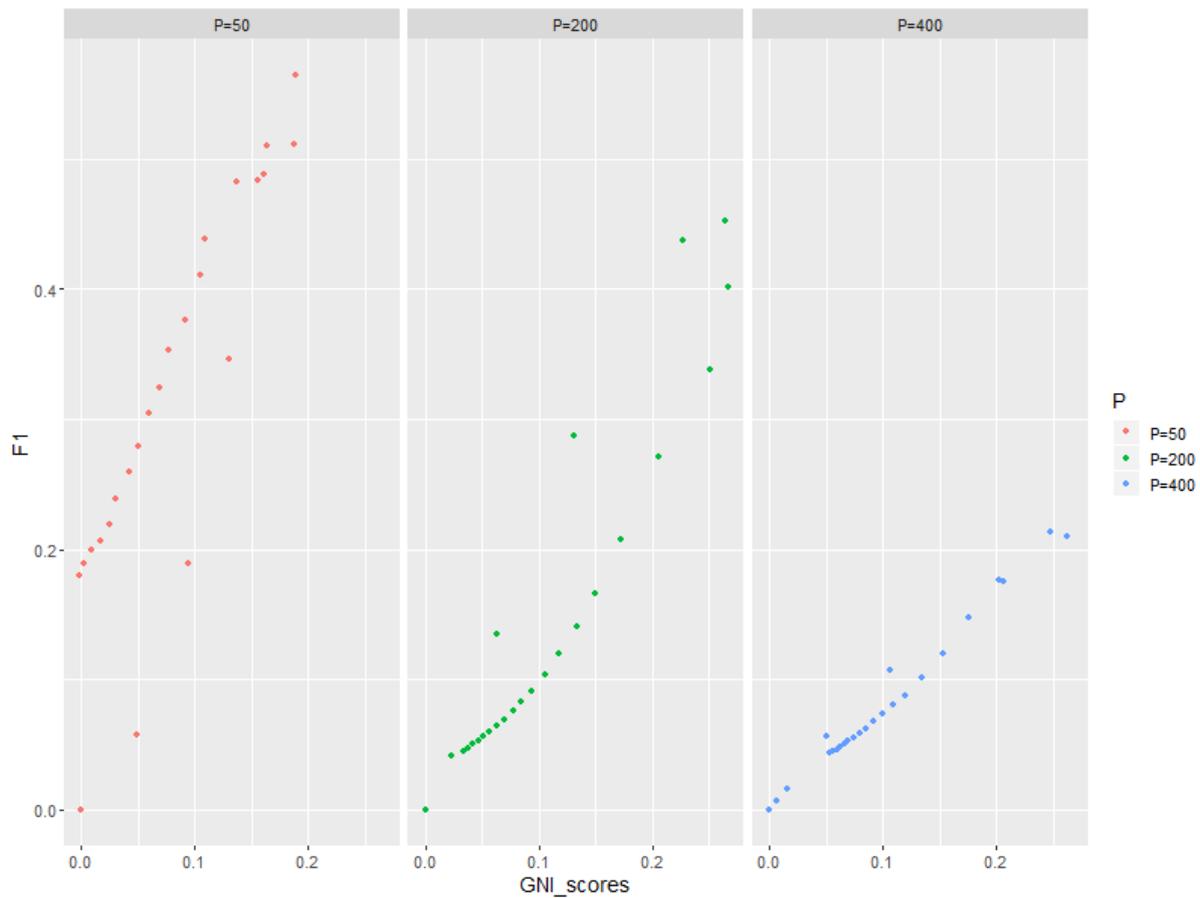

*Figure 6: Visualising the association between GNI scores and F1. Graph inference was performed with GLASSO on simulated data from 3 random graphs with 50, 200 and 400 variables each. Each simulated dataset contained 50 variables. A linear correlation is observed between F1 score and GNI.*

In each dataset, the Graphical Neighbour Information demonstrated a linear correlation with the F1 score of the estimated graph. This correlation was moderate in low-dimensional random graphs (Pearson corr. = 0.766), however it approximated perfect linearity in moderate (Pearson corr. = 0.983) and high dimensional (Pearson corr. = 0.993) random graphs. Strong linear correlation of GNI and F1 was observed in low (Pearson corr. = 0.969), moderate (Pearson corr. = 0.947) and high dimensional (Pearson corr. = 0.897) hub graph simulations.

### Running time

RIC and demonstrated fast running times (50 variables: 0.53s, 200 variables: 0.61s, 400 variables: 0.967s). The true calculation of the EBIC runtime could not be measured as the loglikelihood function was calculated and stored during graph inference by the huge package. It is expected that EBIC runtime would be relatively low. StARS had the longest running time of each method (50 variables: 9.081s, 200 variables: 29.692s, 400 variables: 132.481s). GNI achieved low runtimes (50 variables: 0.443s, 200 variables: 0.817s, 400 variables: 2.063s) in all simulations.

# Discussion

The Graphical Neighbour Information method offers multiple advantages over existing methods for high dimensional graphical model selection.

## Performance

GNI demonstrates clear superiority over all existing methods for model selection. In most cases, GNI selected similarly to the oracle. The gain in performance was greatest in the $n \ll p$ setting, where GNI matched the oracle in terms of F1 and SHD.

## Association of GNI and F1

In our simulations we have observed that GNI shows a high linear correlation with the F1 score of the true graph. This experimentally validates our hypothesis that $\mathbb{P}[\mathcal{H}_w] \propto GNI_{w,\widehat{\Gamma_w}}$. The optimal graph in terms of F1 may be confidently selected as the graph with the highest GNI. This offers an advantage over models with complex selection criteria such those of StARS. Furthermore, an estimate of the performance distribution of the candidate set may be attained, offering insights into the stability of the inference algorithms' results.

## Tuning Free Deployment

Implementation of the Graphical Neighbour Information criterion is simple. The method requires no hyperparameter selection, excepting $m$, the number of inter-observation distances to sample to generate $\boldsymbol{X}^b$. This value may be set at $n^2$ by default. In our experiments we have found that $m \ll n^2$ is sufficient in most situations.

## Computational Cost

Although the RIC and EBIC have low computational cost, they have demonstrated is poor performance in our simulations. StARS offers good performance in the moderately high dimensional setting. However, it requires 25 runs of the GLASSO algorithm on subsamples of the data. This computation proved time-consuming in our simulations, especially in the high dimensional setting. In comparison, GNI demonstrated low running times in all simulations. GNI running tim

## Generalisability

The GNI appraises the fit of an adjacency matrix to a given dataset. Therefore, it may be used to compare sets of adjacency matrices inferred by different algorithms. This offers an advantage over the Rotation Invariance Criterion, which offers only a regularisation parameter and cannot compare multiple models.

## Task Specificity

In many graphical inference tasks, such as biological interactome inference, the primary objective is recovery of the true binary adjacency matrix. In such cases, the priority is the identification of the interactions. Accordingly, the exact values of the pairwise partial correlations are relatively unimportant. The GNI method prioritises the appraisal of the adjacency matrix. Therefore, it is insensitive to model selection errors which may occur due to the fit of the exact values of the non-zero entries of the precision matrix.

## Complete Utilisation of the Training Data

Bootstrapping and cross-validation approaches require the exclusion of a set of observations, which may function as a validation set. The available data for model training is consequently reduced. In the high dimensional setting, this may have a significant impact on the $n\colon p$ ratio. GNI allows full utilisation of the available data.

## Conclusions

Graphical Neighbour Information demonstrates excellent graphical model selection properties in the high dimensional setting. In a given distribution, the model score demonstrates a consistent relationship with the true F1 score of the model. A potential application of the Graphical Neighbour Information criterion is as an objective function for direct gaussian graphical model search.